\pdfoutput=1

\documentclass[11pt]{article}

\usepackage[]{acl}

\usepackage{times}
\usepackage{latexsym}

\usepackage{amsfonts}
\usepackage{graphicx}
\usepackage{bigstrut}
\usepackage{multirow}
\usepackage{booktabs}
\usepackage{amsthm}
\usepackage{mathrsfs}
\usepackage[T1]{fontenc}

\usepackage[utf8]{inputenc}

\usepackage{microtype}

\title{An Accurate Unsupervised Method for Joint Entity Alignment and Dangling Entity Detection}

\author{
Shengxuan Luo$^{1,2}$ \space\space
Sheng Yu$^{1,2}$ \space\space\\
$^{1}$Center for Statistical Science, Tsinghua University \space\space\space\space \\
$^{2}$Department of Industrial Engineering, Tsinghua University \space\space\space\space \\
\texttt{luosx18@mails.tsinghua.edu.cn}\\
\texttt{syu@tsinghua.edu.cn}
}

\usepackage{amsmath}

\begin{document}
\maketitle
\begin{abstract}
Knowledge graph integration typically suffers from the widely existing dangling entities that cannot find alignment cross knowledge graphs (KGs). The dangling entity set is unavailable in most real-world scenarios, and manually mining the entity pairs that consist of entities with the same meaning is labor-consuming. In this paper, we propose a novel accurate \textbf{U}nsupervised method for joint \textbf{E}ntity alignment (EA) and \textbf{D}angling entity detection (DED), called UED. The UED mines the literal semantic information to generate pseudo entity pairs and globally guided alignment information for EA and then utilizes the EA results to assist the DED. We construct a medical cross-lingual knowledge graph dataset, MedED, providing data for both the EA and DED tasks. Extensive experiments demonstrate that in the EA task, UED achieves EA results comparable to those of state-of-the-art supervised EA baselines and outperforms the current state-of-the-art EA methods by combining supervised EA data. For the DED task, UED obtains high-quality results without supervision.
\end{abstract}

\section{Introduction}

Entity alignment (EA) that aligns the equivalent entities in different knowledge graphs (KGs) is a fundamental technique for knowledge graph integration.  A typical application of EA is constructing a large-scale KG by integrating different KGs to facilitate various downstream tasks such as question answering \citep{savenkov2016crqa,yu2017improved,jin2021biomedical}, recommendation \citep{cao2019unifying}, and search engines \citep{xiong2017explicit}. The existing embedding-based EA methods align each entity to its closest counterpart cross KGs according to entity embeddings. In recent years, they have emerged as the dominant EA solutions due to their effectiveness and strong ability to utilize information such as entity name strings, entity description, attributes, and graph structure. These EA methods \citep{chen2017multilingual,sun2018bootstrapping,wang2018cross,zhu2021raga,liu2021self,lin2021echoea} are built upon the assumption that there exists a counterpart in the target KG for any source entity \citep{sun2021knowing}. Therefore, ideally, their performances are assessed by only considering the entities in the set of testing entity pairs.

In the real-world scenario, four facts should be considered when aligning KGs: (1) The entities that do not have counterparts in another KG are ubiquitous. These entities are referred to as dangling entities, following \citet{sun2021knowing}. Therefore, it is necessary to identify the dangling entities and then align the remaining matchable entities to their counterparts. The widely used approach of integrating KGs according to the cross KG similarity between entities loses sight of identifying dangling entities. (2) Dangling entity sets are not labeled in most cases, while some entity pairs are relatively available but labor-consuming. For example, we can preliminarily obtain pseudo entity pairs with high similarity according to extra information to align entities and then manually extract the correct pairs. The extra information could be cross KG links or literal semantic information from machine translation or word embeddings. However, identifying a dangling entity requires manual comparisons between an entity and all entities in the target KG, which is tedious and almost impossible for large KGs. Dangling entity detection (DED) methods need to avoid reliance on supervision. (3) Literal semantic information has an essential impact on EA. As shown in previous works \citep{wu2019relation,nguyen2020entity,zhu2021relation}, competitive EA results can be achieved by translating entity names to the same language and calculating the vector representation from GloVe \citep{pennington2014glove}, suggesting that it is possible to get rid of manually annotated entity pairs by automatically mining literal semantic information. (4) Alignments are associated with each other. Traditional EA methods align entities in the local alignment way by calculating the cross KGs similarity of entities and selecting the most similar entity as EA results. The local alignment neglects the association between alignment and suffers from conflicting many-to-one and many-to-many alignments.

Considering the above facts, we propose UED, an accurate \textbf{U}nsupervised method for joint \textbf{E}A and \textbf{D}ED. For EA, to automatically mine the literal semantic information, we generate pseudo entity pairs for the align loss and design a semantic-based globally guided loss to guide the alignment for all entities, not only for those in entity pairs. For DED, since verifying the dangling entity has to check all the entities in the target KG and the dangling entity set is unavailable, we add empty entities into two KGs and transfer the EA and DED tasks into a modified global optimal transport problem (OTP) to identify dangling entities relying on pseudo entity pairs only. We propose a simple but effective way to reduce the complexity of OTP. Our experiments show that the dangling entity identification mechanism also enhances the EA performance. 

There are several traditional EA datasets widely used in the EA task. Nevertheless, neither dataset provides a dangling test set for DED. As mentioned above, identifying dangling entities is crucial in real-world knowledge graph integration. To demonstrate the effectiveness of our method and incentivize future studies, we construct a cross-lingual medical knowledge graph dataset with EA task and DED task, called MedED, based on the Unified Medical Language System (UMLS) \citep{lindberg1993unified}. 

We summarize the main contributions as follows:
\begin{itemize}
	\item We construct a cross-lingual knowledge graph dataset to demonstrate the effect of our designs and support future studies on EA and DED.
	\item We propose UED, a unified unsupervised method for both EA and DED, which gets rid of supervision in both tasks and fits the real-world scenario when aligning KGs. UED mines the literal semantic information for EA and then utilizes the EA results on pseudo entity pairs to generate high-quality DED results and consequently facilitates the performance of EA.
	\item We conduct comprehensive experiments on both MedED and DBP15K. In the EA task, UED achieves comparable results with state-of-the-art supervised baselines, and the supervised version of UED outperforms the current state-of-the-art methods.
\end{itemize}

The source code of UED is publicly available at \url{https://github.com/luosx18/UED}.

\section{Related Work}
\label{sec:section 2}

\subsection*{Embedding-based Entity Alignment}

Embedding-based entity alignment methods build upon knowledge embedding models, which have been developing rapidly in recent years and aim to encode KGs into low-dimensional vector space. The mainstream embedding-based EA methods adopt models such as TransE \citep{bordes2013translating}, GCN \citep{kipf2016semi}, GAT \citep{velivckovic2017graph}, and the other variants \citep{sun2017cross,zhu2021relation}, to represent entities of different KGs in vector space. Then they find equivalent entity pairs between KGs in the local alignment way. 

The critical point of these EA methods is to include more semantic information in KGs accurately and effectively. The semantic information comprises graph structure, attributes, and literal information, but not all KGs contain all information mentioned above. All embedding-based EA methods adopt graph structures \citep{chen2017multilingual}, while some methods utilize attributes \citep{sun2017cross,trisedya2019entity} or literal information \citep{xu2019cross,wu2019relation,zhu2021raga}. To alleviate the insufficiency of training data, some studies attempt to leverage bootstrapping, iterative training techniques, and self-supervised learning to enrich the training entity pairs with pseudo pairs \citep{sun2018bootstrapping,mao2020relational,liu2021self}.  The proposed method utilizes literal semantic information to generate alignment guidance for all entities in KGs without supervision and is compatible with all graph embedding models mentioned above.

\subsection*{Global Entity Alignment}

Local alignment ignores the fact that alignments are associated with each other, resulting in incorrect alignments and illegal many-to-one and many-to-many alignments \citep{xu2020coordinated,zeng2020collective}. Global EA methods that consider all alignments together have been proposed to mitigate these issues but require relatively good quality local EA to avoid the accumulation of incorrect alignments. Unfortunately, according to the Hungarian algorithm \citep{kuhn1955hungarian}, the complexity of finding the best alignment between two KGs of $n$ entities is $O(n^{4})$. The existing approximate global alignment methods, CEA \citep{zeng2020collective} and GM-EHD-JEA \citep{xu2020coordinated}, reduce the complexity with extra constraints. The CEA requires the entity pairs to be stable matches and uses the deferred acceptance algorithm (DAA) to find the alignments. The GM-EHD-JEA decomposes the entire search space into many isolated subspaces and consequently restricts the cross-subspace alignment. 

\subsection*{Dangling Entity Detection}

Several recent studies emphasize the problem of dangling entities in EA tasks. \citet{zhao2020experimental} and \citet{zeng2021towards} introduce threshold-based methods to identify dangling entities according to the distance between a source entity and its closest target entity. These two methods identify dangling entities to improve EA behavior. \citet{sun2021knowing} also studied the performance of DED in the supervised setting by using the dangling training set to train the classification model or marginal ranking model. 

Our method transfers the global EA and the DED into a modified unified optimal transport problem and consequently relieves the constraints on global EA, utilizes the association between alignment, and does not rely on dangling entity labels.

\section{UED Framework}
\label{sec:section 3}

\begin{figure}[t]
    \centering
    \includegraphics[scale=0.415]{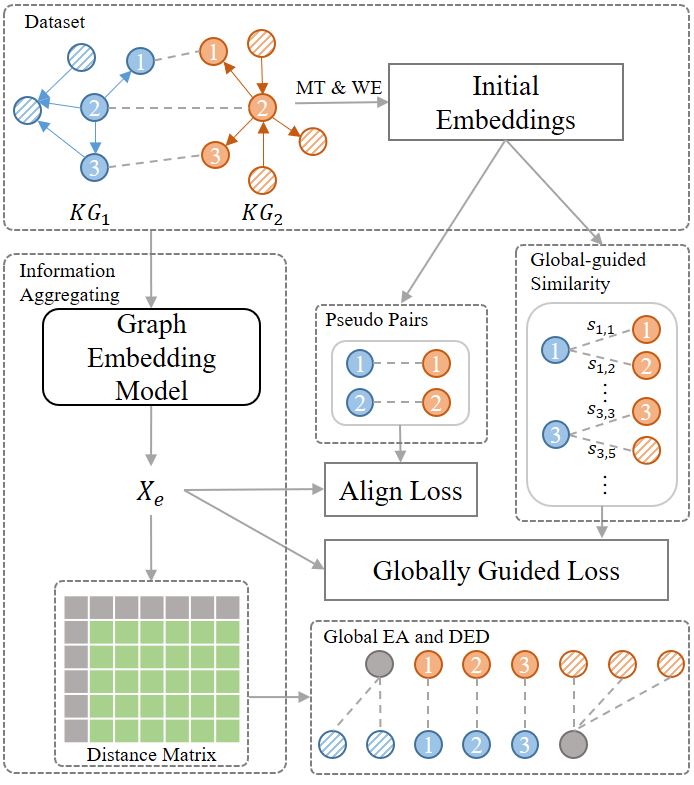}
    \caption{The Framework of UED. The rounded rectangles with dashed line denote the main modules. The circles with a number are matchable entities, and the circles with slash denote dangling entities. The gray circles are the empty entities and the gray rectangles in distance matrix denotes distance between empty entity to other entities. MT and WE refer to machine translation and word embeddings.}
    \label{fig:Figure 1}
\end{figure}

In this section, we first briefly describe the tasks of EA and DED and then elucidate our unified unsupervised approach to solve EA along with DED. An overview of our method is depicted in Figure~\ref{fig:Figure 1}.

\subsection{Task Definition}
Formally, a KG is denoted as $\mathcal{G}=\left\{\mathcal{E},\mathcal{R},\mathcal{T}\right\}$, where $\mathcal{E}=\mathcal{D}\cup\mathcal{A}$ is the disjoint union of dangling set $\mathcal{D}$ and matchable set $\mathcal{A}$. $\mathcal{R}$ and $\mathcal{T}$ denote the set of relations and triples, respectively. For two KGs, $\mathcal{G}_1$ and $\mathcal{G}_2$, the DED task aims to find $\mathcal{D}_1$ and $\mathcal{D}_2$, while the EA task aims to find the entity pairs between the remaining set, $\mathcal{A}_1$ and $\mathcal{A}_2$.

\subsection{Pseudo Entity Pairs}
Manually generating entity pairs to train the embedding base EA model is labor-consuming. We automatically generate pseudo entity pairs for model training, relying only on machine translation and word embeddings.

In our approach, we utilize GloVe \citep{pennington2014glove} word embeddings to generate the mean word vector $v_i$ for entity $e_i$ based on the entity name. Then the initial similarity between $e_i\in\mathcal{G}_1$ and $e_j\in\mathcal{G}_2$ is defined as the cosine similarity $s_{ij} = \cos(v_i, v_j)$. The set of pseudo entity pairs consists of entity pairs with high similarity. Specifically, we define a threshold $\varepsilon<1$. If $s_{ij}$ satisfies: 
\begin{equation}\begin{aligned}
s_{ij}&>\varepsilon,  \\ 
s_{ik}&\leq \varepsilon,\forall k\neq j,  \\
s_{lj}&\leq \varepsilon,\forall l\neq i,
\end{aligned}\end{equation}
then pair $(e_i,e_j)$ is added to the pseudo entity pairs set $\mathcal{P}$. For cross-lingual KGs, we translate entity names using machine translation before applying the word embeddings.

\subsection{Information Aggregating}
\label{sec:section 3.3}
Our method is compatible with all graph embedding models. In this paper, we follow the widespread setting to use relation triples as graph structure information and entity names as literal information \citep{xu2019cross,wu2019relation,mao2020relational,nguyen2020entity,zhu2021raga}. We use a graph embedding model to aggregate the initial embeddings and relation triples to generate enhanced entity embeddings, $X_e$.

Unlike previous works \citep{xu2019cross,mao2020relational,nguyen2020entity,zhu2021raga}, we use pseudo entity pairs to train the graph embedding model instead of training entity pairs. Denoting $X_{e_i}$ as the output embeddings of entity $e_i$ after the graph embedding model, we modify the hinge loss with the pseudo entity pairs, denoted as align loss:
\begin{equation}\begin{aligned}
\mathcal{L}_a = &\sum\limits_{(e_i,e_j)\in\mathcal{P}} \sum\limits_{(e_i^{'},e_j^{'})\in\mathcal{P}^{'}(e_i,e_j)} \max\Big(d\left(X_{e_i},X_{e_j}\right)\\ &\left.-d\left(X_{e_i^{'}},X_{e_j^{'}}\right)+\lambda,0\right),
\end{aligned}\end{equation}
where $\lambda$ is the margin, $\mathcal{P}^{'}(e_i,e_j)$ is the set of negative samples for $(e_i, e_j)$ by replacing $e_i$ or $e_j$ with their neighbors, and $d(\cdot,\cdot)$ is the Manhattan distance following previous works \citep{wu2019relation,zhu2021raga}.

\subsection{Globally Guided Similarity and Loss}
The align loss does not make full use of literal semantic information since the initial similarity $s_{ij}$ contains entity alignment information for entities not in $\mathcal{P}$. In addition, training an EA model with the align loss may mislead the model to pay too much attention to the entities in $\mathcal{P}$. Therefore, we regard entities in the target KG as anchors to guide the EA training for all source entities. Our assumption is that the counterpart of an entity is more likely to occur among entities whose initial embeddings are more similar. Specifically, we propose a globally guided loss:
\begin{equation}\begin{aligned}
\mathcal{L}_g = &\sum\limits_{(e_i,e_j)\in\mathcal{Q}} s_{ij}\sum\limits_{(e_i^{'},e_j^{'})\in\mathcal{Q}^{'}(e_i,e_j)} \max\Big(d\left(X_{e_i},X_{e_j}\right)\\ &\left.-d\left(X_{e_i^{'}},X_{e_j^{'}}\right)+\lambda,0\right),
\end{aligned}\end{equation}
where $\mathcal{Q}$ consists of all $(e_i, e_j)$ satisfying $e_j$ is one of the top $k$ similar entities of $e_i$ according to the initial semantic similarity $\{s_{ij},\forall j\}$, and $k$ is a hyperparameter. The construction of $\mathcal{Q}^{'}$ is similar to $\mathcal{P}^{'}$. According to our experiments, $s_{ij}$ is a necessary value that refers to the weight of $(e_i, e_j)$ in $\mathcal{L}_g$ to improve model performance. To gradually reduce the impact of entities in $\mathcal{Q}$, we design a mechanism to decrease the weight of the globally guided loss. The final loss is
\begin{equation}
\mathcal{L} = \mathcal{L}_a+w(t)\mathcal{L}_g,
\end{equation} 
where $t$ is the training step, and $w(t)$ decreases linearly to 0 as $t$ increases.

\subsection{Global EA and DED}
Given two KGs comprising $n$ and $m$ entities, we define a distance matrix $C\in {\mathbb{R}} ^{n\times m}$ with each entry indicating the Manhattan distance between two entities. The global EA task can be formulated into an optimal transport problem (OTP) to find an optimal global alignment by minimizing the total transport distance:
\begin{equation}\begin{aligned}
\min &\sum\limits_{i=1,j=1}\limits^{n,m}C_{ij}\Psi_{ij},\\
\operatorname{s.t. } &\sum\limits_j\Psi_{ij}=1,1\leq i\leq n,\\
&\sum\limits_i\Psi_{ij}=1,1\leq j\leq m,
\end{aligned}\end{equation} 
where $\Psi$ is the transport matrix, and $\Psi_{ij}\in\{0,1\}$ for all $i$ and $j$ indicates whether entity $e_i$ in $\mathcal{G}_1$ aligns to $e_j$ in $\mathcal{G}_2$. The constraints guarantee the one-to-one alignment. Considering that $n\neq m$ in most cases and the existence of dangling entities, this OTP is invalid. To address these issues, we add an empty entity into $\mathcal{G}_1$ and $\mathcal{G}_2$ separately. Without loss of generality, we prepend the empty entity as the first entity in both KGs. Since we have no information for empty entities, we define hyperparameters, $\alpha$ and $\beta$, to describe the cross KG distance between the empty entity and other entities. Therefore, the OTP is now as follow:
\begin{equation}\begin{aligned}
\min &\sum\limits_{i=1,j=1}\limits^{n+1,m+1}C_{ij}\Psi_{ij},\\
\operatorname{s.t. } &\sum\limits_j\Psi_{ij}=1,2\leq i\leq n+1,\\
&\sum\limits_i\Psi_{ij}=1,2\leq j\leq m+1,
\end{aligned}\end{equation} 
where $C_{1,j}=\alpha,\forall j$ and $C_{i,1}=\beta,\forall i$ denote the first row and the first column of the distance matrix, respectively. $\Psi_{ij}\in\{0,1\}$, and $\Psi_{i,1}=1$ indicates that entity $e_i$ is dangling, while $\Psi_{1,j}=1$ also indicates dangling entity $e_j$. The other $\Psi_{i,j}=1$ predicts the entity pair $(e_i,e_j)$.

Our approach now merges the EA and the DED into one OTP. This OTP considers the global alignment information and the interactions among alignments and dangling entity identification. Moreover, considering that similar entities contain more information for both EA and DED, we keep the top $K$ rank similarity entities in the other KG for each entity and drop the remaining entities to reduce the complexity of the OTP. Therefore, we solve the problem with very sparse matrices, $C$ and $\Psi$. Section~\ref{sec:section 5.3} will show that the method is powerful with acceptable computational complexity after reduction. The last problem is to find the proper $\alpha$ and $\beta$ for both EA and DED. Since we have the pseudo entity pairs set $\mathcal{P}$ in real-world data, we propose an ingenious way to grid search the quantiles of row minimums and column minimums of $C$ synchronously and then select $\alpha^*$ and $\beta^*$ that achieve the best EA performance on $\mathcal{P}$. Finally, the entities aligned to the empty entity under given $\alpha^*$ and $\beta^*$ are dangling entities. The other alignments are the global EA results. 

\section{Experimental Setup}
\subsection{Datasets and Evaluation}
\citet{sun2021knowing} construct a dataset providing EA task and DED task, which contains the information of relation triples only so that the quality of local EA is limited and therefore incompatible with global alignment methods. In this work, we construct a dataset with graph structure and literal semantic information providing both EA and DED tasks.

\begin{table}[t]
\setlength\tabcolsep{2pt}
\small
  \centering
    \begin{tabular}{ccccccc}
    \hline
    \multicolumn{2}{c}{Datasets} &
      \multicolumn{1}{c}{\#Ent.} &
      \multicolumn{1}{c}{\#Rel.} &
      \multicolumn{1}{c}{\#Trip.} &
      \multicolumn{1}{c}{\#Pairs} &
      \#Dang.
      \bigstrut\\
    \hline
    \multicolumn{1}{c}{\multirow{2}[1]{*}{MedED}} &
      FR &
      19,382 &
      431 &
      455,368 &
      \multirow{2}[1]{*}{6,365} &
      \multicolumn{1}{c}{13,017}
      \bigstrut[t]\\
     &
      EN &
      18,632 &
      622 &
      841,792 &
       &
      \multicolumn{1}{c}{12,267}
      \\
    \multicolumn{1}{c}{\multirow{2}[0]{*}{MedED}} &
      ES &
      19,228 &
      546 &
      594,130 &
      \multirow{2}[0]{*}{11,153} &
      \multicolumn{1}{c}{8,075}
      \\
     &
      EN &
      18,632 &
      622 &
      841,792 &
       &
      \multicolumn{1}{c}{7,479}
      \\
    \multicolumn{1}{c}{\multirow{2}[0]{*}{DBP15K}} &
      ZH &
      19,388 &
      1,700 &
      70,414 &
      \multirow{2}[0]{*}{15,000} &
      -
      \\
     &
      EN &
      19,572 &
      1,322 &
      95,142 &
       &
      -
      \\
    \multicolumn{1}{c}{\multirow{2}[0]{*}{DBP15K}} &
      JA &
      19,814 &
      1,298 &
      77,214 &
      \multirow{2}[0]{*}{15,000} &
      -
      \\
     &
      EN &
      19,780 &
      1,152 &
      93,484 &
       &
      -
      \\
    \multicolumn{1}{c}{\multirow{2}[1]{*}{DBP15K}} &
      FR &
      19,661 &
      902 &
      105,998 &
      \multirow{2}[1]{*}{15,000} &
      -
      \\
     &
      EN &
      19,993 &
      1,207 &
      115,722 &
       &
      -
      \bigstrut[b]\\
    \hline
    \end{tabular}%
    \caption{Statistics of MedED and DBP15K.}
  \label{tab:Table 1}%
\end{table}%

\subsubsection*{Dataset Construction}
The Unified Medical Language System (UMLS) \citep{lindberg1993unified} is a large-scale resource containing over 4 million unique medical concepts and over 87 million relation triples. Concepts in UMLS have several terms in different languages. We extract concepts that contain terms in the selected language as entities to construct new monolingual KG and retain the relations between entities. For the entity names, we select the preferred terms in UMLS. The criterion of entity pairs is whether entities belong to the same concept. Similarly, an entity is dangling if its original concept is not in the other KG. We extracted the KGs of English, French, and Spanish and then constructed the KG pairs of FR-EN (French to English) and ES-EN (Spanish to English). We select 20 thousand entities with the most relation triples in UMLS for the specified language and then drop the entities unrelated to other selected entities. Table~\ref{tab:Table 1} shows the statistics of the new dataset, MedED. For both EA and DED, we split 70\% of entity pairs and dangling entities as the test set. Even though our method does not rely on the training set, we keep the remaining 30\% as the training set for further model comparison and ablation study.

\subsubsection*{DBP15K}
We conduct experiments on the widely used existing EA benchmark, DBP15K \citep{sun2017cross}. Three pairs of cross-lingual KGs, ZH-EN (Chinese to English), JA-EN (Japanese to English), and FR-EN (French to English), were built into this dataset. Each KG contains approximately 20 thousand entities, and every KG pair contains 15 thousand entity pairs (Table~\ref{tab:Table 1}). Following the setting in previous works \citep{sun2017cross,wu2019relation,zhu2021raga}, we keep 70\% of entity pairs for testing and 30\% for training.
\subsubsection*{Evaluation}
We compute two evaluation metrics following previous works for the EA task, Hits@k and mean reciprocal rank (MRR). Hits@k indicates the percentage of the targets that have been correctly ranked in the top K. MRR is the average of the reciprocal of the rank results. The previous EA works compute Hits@k and MRR in a \emph{relaxed setting} in which only the entities in testing pairs are taken into account, assuming that any source entity has a counterpart in the target KG. In addition to the relaxed evaluation, we also compute Hits@k and MRR in a \emph{practical setting} in which for every testing entity, the list of candidate counterparts consists of all entities in the other KG. Global alignment methods generate one-to-one entity pairs, and we evaluate Hits@1 for these methods.

For the DED task, we compute precision, recall, and F1-score for identifying dangling entities.

\subsection{Compared Methods}
For the EA task, we compare our approach with previous methods we introduced in Section~\ref{sec:section 2}: (1) Init-Emb, the initial embeddings used in UED and main comparison models; (2) the methods based on translational KG embeddings model: MTransE \citep{chen2017multilingual}, JAPE \citep{sun2017cross}, and BootEA \citep{sun2018bootstrapping}; (3) the methods based on graph neural networks: RDGCN \citep{wu2019relation}, CEA \citep{zeng2020collective}, RNM \citep{zhu2021relation}, RAGA \citep{zhu2021raga}, SelfKG \citep{liu2021self}, EchoEA \citep{lin2021echoea}. 

The proposed method is compatible with supervised training entity pairs, so we provide both unsupervised and supervised versions of our method: (1) the unsupervised method, UED, described in Section~\ref{sec:section 3}. (2) the supervised version of UED, which combines the training entity pairs and the pseudo entity pairs for the align loss, denoted as UED*. 

\subsection{Implementation Details}
Following \citet{wu2019relation}, we translate entity names in MedED to English via Google Translate and then use mean of word vector from GloVe \citep{pennington2014glove} to construct the initial entity embeddings. For entities in DBP15K, we inherit the initial embeddings used in previous works \citep{wu2019relation,zeng2021towards,zhu2021raga,zhu2021relation,lin2021echoea}. The threshold for pseudo entity pairs $\varepsilon$ is 0.99, and the $k=3$ in globally guided similarity and loss. The initial value of $w(t)$ is 0.3 and $w(t)$ decreases linearly to 0 at 1/4 of the total training steps. We adopt RAGA \citep{zhu2021raga} as the embedding-based EA model in Section~\ref{sec:section 3.3} to generate enhanced entity embeddings and use the default setting of hyperparameters in RAGA. For $\alpha^*$ and $\beta^*$  in the global EA and DED, the default value of $K$ is 100 for our method. We grid search 100  paired quantiles of the row minimums and column minimums of $C$ with $K=10$. Then, $\alpha^*$ and $\beta^*$ are used in the other values of $K$.

\begin{table*}[t]
  \setlength\tabcolsep{2.2pt}
  \small
  \centering
    \begin{tabular}{lccccccccccccccc}
    \toprule
     &
      \multicolumn{9}{c}{DBP15K} &
      \multicolumn{6}{c}{MedED}
      \\
    \midrule
     &
      \multicolumn{3}{c}{ZH-EN} &
      \multicolumn{3}{c}{JA-EN} &
      \multicolumn{3}{c}{FR-EN} &
      \multicolumn{3}{c}{FR-EN} &
      \multicolumn{3}{c}{ES-EN}
      \\
     &
      H@1 &
      \multicolumn{1}{c}{H@10} &
      \multicolumn{1}{c}{MRR} &
      H@1 &
      \multicolumn{1}{c}{H@10} &
      \multicolumn{1}{c}{MRR} &
      H@1 &
      \multicolumn{1}{c}{H@10} &
      \multicolumn{1}{c}{MRR} &
      H@1 &
      \multicolumn{1}{c}{H@10} &
      \multicolumn{1}{c}{MRR} &
      H@1 &
      \multicolumn{1}{c}{H@10} &
      \multicolumn{1}{c}{MRR}
      \\
    \midrule
    \textbf{Local } &
       &
       &
       &
       &
       &
       &
       &
       &
       &
       &
       &
       &
       &
       &
      
      \\
    \midrule
    \underline{Init-Emb} &
      .575  &
      \multicolumn{1}{c}{.689 } &
      \multicolumn{1}{c}{.615 } &
      .650  &
      \multicolumn{1}{c}{.754 } &
      \multicolumn{1}{c}{.688 } &
      .818  &
      \multicolumn{1}{c}{.888 } &
      \multicolumn{1}{c}{.843 } &
      .716  &
      \multicolumn{1}{c}{.845 } &
      \multicolumn{1}{c}{.764 } &
      .685  &
      \multicolumn{1}{c}{.826 } &
      \multicolumn{1}{c}{.737 }
      \\
    MTransE &
      .308  &
      \multicolumn{1}{c}{.614 } &
      \multicolumn{1}{c}{.364 } &
      .279  &
      \multicolumn{1}{c}{.575 } &
      \multicolumn{1}{c}{.349 } &
      .244  &
      \multicolumn{1}{c}{.556 } &
      \multicolumn{1}{c}{.335 } &
      - &
      \multicolumn{1}{c}{-} &
      \multicolumn{1}{c}{-} &
      - &
      \multicolumn{1}{c}{-} &
      \multicolumn{1}{c}{-}
      \\
    JAPE &
      .731  &
      \multicolumn{1}{c}{.904 } &
      \multicolumn{1}{c}{-} &
      .828  &
      \multicolumn{1}{c}{.947 } &
      \multicolumn{1}{c}{-} &
      - &
      \multicolumn{1}{c}{-} &
      \multicolumn{1}{c}{-} &
      - &
      \multicolumn{1}{c}{-} &
      \multicolumn{1}{c}{-} &
      - &
      \multicolumn{1}{c}{-} &
      \multicolumn{1}{c}{-}
      \\
    BootEA &
      .629  &
      \multicolumn{1}{c}{.848 } &
      \multicolumn{1}{c}{.703 } &
      .622  &
      \multicolumn{1}{c}{.854 } &
      \multicolumn{1}{c}{.701 } &
      .653  &
      \multicolumn{1}{c}{.874 } &
      \multicolumn{1}{c}{.731 } &
      - &
      \multicolumn{1}{c}{-} &
      \multicolumn{1}{c}{-} &
      - &
      \multicolumn{1}{c}{-} &
      \multicolumn{1}{c}{-}
      \\
    \underline{RDGCN} &
      .708  &
      \multicolumn{1}{c}{.846 } &
      \multicolumn{1}{c}{-} &
      .767  &
      \multicolumn{1}{c}{.895 } &
      \multicolumn{1}{c}{-} &
      .886  &
      \multicolumn{1}{c}{.957 } &
      \multicolumn{1}{c}{-} &
      - &
      \multicolumn{1}{c}{-} &
      \multicolumn{1}{c}{-} &
      - &
      \multicolumn{1}{c}{-} &
      \multicolumn{1}{c}{-}
      \\
    \underline{RNM} &
      .840  &
      \multicolumn{1}{c}{.919 } &
      \multicolumn{1}{c}{\textbf{.870 }} &
      .872  &
      \multicolumn{1}{c}{.944 } &
      \multicolumn{1}{c}{.899 } &
      .938  &
      \multicolumn{1}{c}{.981 } &
      \multicolumn{1}{c}{.954 } &
       &
      \multicolumn{1}{c}{-} &
      \multicolumn{1}{c}{-} &
      - &
      \multicolumn{1}{c}{-} &
      \multicolumn{1}{c}{-}
      \\
    \underline{RAGA} &
      .798  &
      \multicolumn{1}{c}{.930 } &
      \multicolumn{1}{c}{.847 } &
      .831  &
      \multicolumn{1}{c}{.950 } &
      \multicolumn{1}{c}{.875 } &
      .914  &
      \multicolumn{1}{c}{.983 } &
      \multicolumn{1}{c}{.940 } &
      .896  &
      \multicolumn{1}{c}{\textbf{.981 }} &
      \multicolumn{1}{c}{.930 } &
      \textbf{.914 } &
      \multicolumn{1}{c}{.986 } &
      \multicolumn{1}{c}{\textbf{.943 }}
      \\
    \underline{SelfKG} &
      .829  &
      \multicolumn{1}{c}{.919 } &
      \multicolumn{1}{c}{-} &
      .890  &
      \multicolumn{1}{c}{.953 } &
      \multicolumn{1}{c}{-} &
      .959  &
      \multicolumn{1}{c}{\textbf{.992 }} &
      \multicolumn{1}{c}{-} &
      - &
      \multicolumn{1}{c}{\textbf{-}} &
      \multicolumn{1}{c}{-} &
      \textbf{-} &
      \multicolumn{1}{c}{-} &
      \multicolumn{1}{c}{\textbf{-}}
      \\
    \underline{EchoEA} &
      .823  &
      \multicolumn{1}{c}{.939 } &
      \multicolumn{1}{c}{.865 } &
      .861  &
      \multicolumn{1}{c}{.957 } &
      \multicolumn{1}{c}{.897 } &
      .939  &
      \multicolumn{1}{c}{.989 } &
      \multicolumn{1}{c}{.958 } &
      - &
      \multicolumn{1}{c}{\textbf{-}} &
      \multicolumn{1}{c}{-} &
      \textbf{-} &
      \multicolumn{1}{c}{-} &
      \multicolumn{1}{c}{\textbf{-}}
      \\
    \underline{UED} &
      .779  &
      \multicolumn{1}{c}{.907 } &
      \multicolumn{1}{c}{.826 } &
      .820  &
      \multicolumn{1}{c}{.933 } &
      \multicolumn{1}{c}{.862 } &
      .921  &
      \multicolumn{1}{c}{.979 } &
      \multicolumn{1}{c}{.943 } &
      .895  &
      \multicolumn{1}{c}{.975 } &
      \multicolumn{1}{c}{.926 } &
      .893  &
      \multicolumn{1}{c}{.978 } &
      \multicolumn{1}{c}{.925 }
      \\
    \underline{UED*} &
      .826  &
      \multicolumn{1}{c}{\textbf{.943 }} &
      \multicolumn{1}{c}{\textbf{.870 }} &
      .863  &
      \multicolumn{1}{c}{.960 } &
      \multicolumn{1}{c}{\textbf{.900 }} &
      .938  &
      \multicolumn{1}{c}{.987 } &
      \multicolumn{1}{c}{\textbf{.957 }} &
      .901  &
      \multicolumn{1}{c}{\textbf{.981 }} &
      \multicolumn{1}{c}{\textbf{.932 }} &
      .913  &
      \multicolumn{1}{c}{\textbf{.987 }} &
      \multicolumn{1}{c}{.942 }
      \\
    \midrule
    \textbf{Global} &
       &
       &
       &
       &
       &
       &
       &
       &
       &
       &
       &
       &
       &
       &
      
      \\
    \midrule
    GM-EHD-JEA &
      .736  &
       &
       &
      .792  &
       &
       &
      .924  &
       &
       &
      - &
       &
       &
      - &
       &
      
      \\
    CEA &
      .787  &
       &
       &
      .863  &
       &
       &
      .972  &
       &
       &
      - &
       &
       &
      - &
       &
      
      \\
    \underline{RAGA} &
      .873  &
       &
       &
      .909  &
       &
       &
      .966  &
       &
       &
      .962  &
       &
       &
      .970  &
       &
      
      \\
    \underline{EchoEA} &
      .891  &
       &
       &
      .932  &
       &
       &
      \textbf{.989} &
       &
       &
      - &
       &
       &
      - &
       &
      
      \\
    \underline{UED} &
      .877  &
       &
       &
      .915  &
       &
       &
      .975  &
       &
       &
      .970  &
       &
       &
      .976  &
       &
      
      \\
    \underline{UED*} &
      \textbf{.915} &
       &
       &
      \textbf{.941} &
       &
       &
      .984  &
       &
       &
      \textbf{.974} &
       &
       &
      \textbf{.979} &
       &
      
      \\
    \bottomrule
    \end{tabular}%
  \caption{EA results on DBP15K and MedED datasets (relaxed setting). H@1 and H@10 denotes the Hits@1 and Hits@10. The underlined models use the same initial entity embeddings. The results of the compared method in DBP15K are from their original papers. We apply the RAGA in MedED for comparison. The CEA, RAGA and EchoEA use the DAA for global alignment.}
  \label{tab:Table 2}%
\end{table*}%

\begin{table}[t]
  \setlength\tabcolsep{2pt}
  \small
  \centering
    \begin{tabular}{lcccccccc}
    \toprule
     &
      \multicolumn{4}{c}{FR-EN} &
      \multicolumn{4}{c}{ES-EN}
      \\
    \midrule
     &
      EA &
      \multicolumn{3}{c}{DED} &
      EA &
      \multicolumn{3}{c}{DED}
      \\
     &
      H@1 &
      P &
      R &
      F &
      H@1 &
      P &
      R &
      F
      \\
    \midrule
    RAGA &
      .787  &
      - &
      - &
      - &
      .827  &
      - &
      - &
      -
      \\
    UED(DAA) &
      .774  &
      - &
      - &
      - &
      .870  &
      - &
      - &
      -
      \\
    Distance &
      - &
      .781  &
      .734  &
      .757  &
      - &
      .786  &
      \textbf{.861} &
      .822 
      \\
    UED &
       &
       &
       &
       &
       &
       &
       &
      
      \\
    {\ \ \ \ \ K=1} &
      .798  &
      .961  &
      \textbf{.794} &
      \textbf{.869} &
      .860  &
      .904  &
      .842  &
      \textbf{.872}
      \\
    {\ \ \ \ \ K=10} &
      .803  &
      .963  &
      .753  &
      .845  &
      .874  &
      .935  &
      .684  &
      .790 
      \\
    {\ \ \ \ \ K=100} &
      .805  &
      .964  &
      .748  &
      .842  &
      .877  &
      .933  &
      .646  &
      .764 
      \\
    UED* &
      \textbf{.826} &
      \textbf{.976} &
      .654  &
      .783  &
      \textbf{.901} &
      \textbf{.941} &
      .694  &
      .799 
      \\
    \bottomrule
    \end{tabular}%
  \caption{EA and DED results on MedED (practical setting). H@1, P, R, and F denotes Hits@1, precision, recall, and F-score. $K = 1,10,100$ refers to the proposed global alignment method that keeps the top $K (= 1, 10, 100)$ rank similarity entities for each entity. The UED(DAA) and RAGA use the DAA for global alignment.}
  \label{tab:Table 3}%
\end{table}%

\section{Results}
\subsection{Entity Alignment Results}
Table~\ref{tab:Table 2} shows the results of EA on DBP15K and MedED. Following the previous work, we adopt the relaxed evaluation setting. The results with practical evaluation setting are listed in Appendix~\ref{sec:appendixA1}.

In general, for both local and global alignment in DBP15K, the UED achieves comparable results with the previous state-of-the-art baselines. More specifically, for local alignment, the UED achieves the same level behavior as the supervised embedding-based EA method, the RAGA, of which we adopt its graph embedding models. For global alignment, the OTP brings UED a significant improvement, and the UED outperforms all competing methods except the new supervised state-of-the-art method, EchoEA. The Hits@1 of UED for ZH-EN, JA-EN, and FR-EN achieves 0.877, 0.915, and 0.975 in DBP15K, respectively. In addition, UED* outperforms all methods and achieves 0.915 and 0.941 Hits@1 for ZH-EN and JA-EN in DBP15K and 0.974 and 0.979 for FR-EN and ES-EN in MedED.

\subsection{Entity Alignment and Dangling Entity Detection Results}
Table~\ref{tab:Table 3} shows the results of EA and DED on MedED. Note that global alignment with DED should consider all entities. We select the practical setting in the EA evaluation. 

As shown in Table~\ref{tab:Table 3}, for the EA task, by maximizing the performance of EA on pseudo entity pairs, UED achieves better results compared to the supervised RAGA and the variants of our method with DAA. In addition, the UED ($K=100$) achieves 0.805 and 0.877 Hits@1 for FR-EN and ES-EN separately. The supervised UED* gains a further improvement of 0.021 and 0.012 Hits@1 for FR-EN and ES-EN separately. For the DED task, the proposed method focuses more on the precision in recognizing dangling entities. The results of UED and UED* are also much better than the Distance. The Distance denotes the baseline by searching the best threshold on the dangling training set for identifying dangling entities according to the smallest distance to entities in another KG. These results imply that UED successfully uses unsupervised EA to assist DED while DED with high precision reduces the scope of EA and enhances the performance of EA.  Furthermore, the results with different $K$ show that we don't need a vary large value of $K$, and there is a tradeoff between improving EA results and DED results:  the larger $K$ achieves the better Hits@1 in the EA task and precision in the DED task, while the smaller $K$ achieves the better F1-score in the DED task. 

\subsection{Empirical Runtime Analysis}
\label{sec:section 5.3}

The time complexity of the proposed global method is acceptable. The solving process of the OTP could be finished in less than 7, 60, and 5,00 seconds for $K=1,10,100$ in MedED. Without the simplification, the running time will be more than 120,000 seconds. Considering the time consuming and the similar performance of $K=10$ and $K=100$ (Table~\ref{tab:Table 3}), much larger value of $K$ may not bring significant improvement and $K=100$ is enough for the proposed method. 

\section{Ablation Study}
To quantify the role of our designs, we provide the variants by removing the weight decreasing mechanism of the globally guided loss $\mathcal{L}_g$ and the $\mathcal{L}_g$ from UED (Table~\ref{tab:Table 4}). In addition, we attempt to replace the proposed OTP with DAA (Table~\ref{tab:Table 4}). For local alignment, the UED without $\mathcal{L}_g$ is the same as RAGA except for the training entity pairs. Table~\ref{tab:Table 5} provides other necessary results and variants in practical setting.  There are five major observations:

\begin{table}[t]
  \setlength\tabcolsep{5pt}
  \small
  \centering
    \begin{tabular}{lccccc}
    \toprule
     &
      \multicolumn{3}{c}{DBP15K} &
      \multicolumn{2}{c}{MedED}
      \\
     &
      ZH &
      JA &
      FR &
      FR &
      ES
      \\
    \midrule
    Local &
       &
       &
       &
       &
      
      \\
    \midrule
    RAGA &
      .798  &
      .831  &
      .914  &
      .896  &
      .914 
      \\
    UED &
      .779  &
      .820  &
      .929  &
      .895  &
      .893 
      \\
    {\ \ \ \ \ w/o $\mathcal{L}_g$}  &
      .759  &
      .794  &
      .913  &
      .891  &
      .896 
      \\
    \midrule
    Global  &
       &
       &
       &
       &
      
      \\
    \midrule
    UED &
      .877  &
      .915  &
      .975  &
      .970  &
      .976 
      \\
    {\ \ \ \ \ w/o dec.} &
      .873  &
      .910  &
      .973  &
      .969  &
      .973 
      \\
    {\ \ \ \ \ w/o $\mathcal{L}_g$} &
      .875  &
      .910  &
      .973  &
      .971  &
      .975 
      \\
    {\ \ \ \ \ w/o OTP} &
      .779  &
      .820  &
      .921  &
      .895  &
      .893 
      \\
    UED(DAA) &
      .847  &
      .891  &
      .962  &
      .955  &
      .956 
      \\
    \bottomrule
    \end{tabular}%
  \caption{Hits@1 results of method variants (relaxed setting) in the EA task. The dec. is the weight decreasing mechanism of the globally guided loss, $\mathcal{L}_g$. ZH, JA, FR and ES denotes the KG pairs ZH-EN, JA-EN, FR-EN and ES-EN.}
  \label{tab:Table 4}%
\end{table}%

\begin{table}[t]
  \centering
  \small
    \begin{tabular}{lcccc}
    \toprule
     &
      \multicolumn{2}{c}{      FR-EN} &
      \multicolumn{2}{c}{     ES-EN}
      \\
     &
      EA &
      DED &
      EA &
      DED
      \\
    \midrule
    UED &
      .803 &
      .845 &
      .874 &
      .790
      \\
    {\ \ \ \ \ w/o empty} &
      .555 &
      - &
      .652 &
      -
      \\
    {\ \ \ \ \ w. gold $\alpha, \beta$} &
      .809 &
      .803 &
      .874 &
      .790
      \\
    UED(CODER) &
      .884  &
      .863  &
      .933  &
      .865 
      \\
    \bottomrule
    \end{tabular}%
  \caption{Results of method variants (practical setting) in MedED. We report Hits@1 and F-score for EA and DED. The w/o empty denotes the OTP without the empty entities. The w. gold $\alpha, \beta$ denote that  the $\alpha$ and $\beta$ in the OTP are selected by the dangling training set. UED(CODER) refers to the method that we replace the Glove with a medical language model in UED.}
  \label{tab:Table 5}%
\end{table}%

\begin{figure}[t]
    \centering
    \includegraphics[scale=0.55]{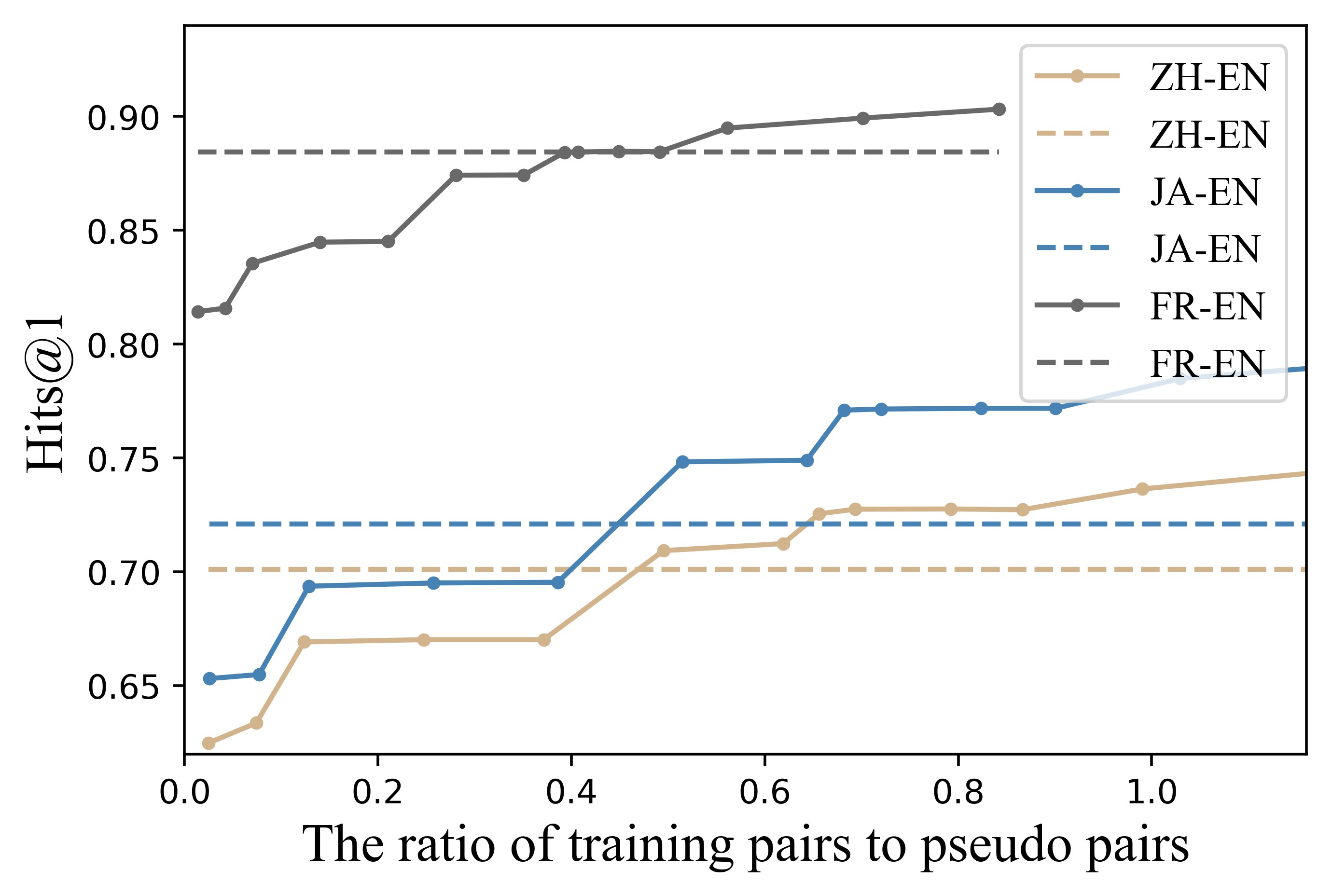}
    \caption{The Hits@1 in DBP15K (practical setting) for the UED without $\mathcal{L}_g$ and OTP. The solid line and dashed line denotes the method trained with the training entity pairs and pseudo entity pairs, respectively. }
    \label{fig:Figure 2}
\end{figure}

1. The performance of our method with pseudo entity pairs is similar to those with true entity pairs. For example, in Table~\ref{tab:Table 4}, for local alignment results of FR-EN in DBP15K, the UED without $\mathcal{L}_g$ uses 10,689 pseudo entity pairs and gains 0.913 Hits@1, while the RAGA uses 4500 true entity pairs and gains 0.914 Hits@1. Although the proportion of how many pseudo entity pairs can play an equal role as true entity pairs changes, depending on the quality of the initial entity embedding and the KGs (Figure ~\ref{fig:Figure 2}), it is valid to obtain pseudo entity pairs when true entity pairs are unavailable.

2. The proposed global alignment method is stable and effective, causing significant improvements (0.046$\sim$0.098 Hits@1) compared with the UED for local alignment Table~\ref{tab:Table 4}). 

3. The globally guided similarity and loss and the weight decreasing mechanism are usually helpful (Table~\ref{tab:Table 4}).

4. Introducing the empty entity is necessary. The global method without empty entities harms the EA result and cannot be applied to the DAD task (Table~\ref{tab:Table 5}). 

5. The proposed method for searching proper $\alpha^*$ and $\beta^*$ produces successful results. The results with $\alpha^*$ and $\beta^*$ achieve the same level of performance for EA and DED compared to the gold selection for $\alpha$ and $\beta$ based on the EA training entity pairs.

Besides, we attempt to replace the GloVe in MedED with a pretrained medical language model (LM), the English version of CODER \citep{yuan2020coder}, and show that a proper domain-specific LM trained on a large KG may achieve better results (Table~\ref{tab:Table 5}).

\section{Conclusion}
This paper proposes a novel unified unsupervised method for both EA and DED, which better fits the realistic scenario for integrating KGs. UED contains four modules: pseudo entity pair generation, information aggregation, globally guided similarity and loss, and a modified OTP for global EA and DED. The first three modules mine the information in KGs to get rid of supervised entity pairs, while the last module integrates EA and DED into a unified framework to identify dangling entities without supervision and provide better EA results. We also construct a new dataset for the EA and DED tasks and perform experiments to demonstrate the effectiveness of UED.

\section*{Acknowledgements}
We would like to express our gratitude to the reviewers for their helpful comments and suggestions. We thank Zheng Yuan, Hongyi Yuan, Pengyu Cheng, and Huaiyuan Ying for their help.

\bibliography{anthology,custom}

\begin{thebibliography}{30}
\expandafter\ifx\csname natexlab\endcsname\relax\def\natexlab#1{#1}\fi

\bibitem[{Bordes et~al.(2013)Bordes, Usunier, Garcia-Duran, Weston, and
  Yakhnenko}]{bordes2013translating}
Antoine Bordes, Nicolas Usunier, Alberto Garcia-Duran, Jason Weston, and Oksana
  Yakhnenko. 2013.
\newblock Translating embeddings for modeling multi-relational data.
\newblock \emph{Advances in neural information processing systems}, 26.

\bibitem[{Cao et~al.(2019)Cao, Wang, He, Hu, and Chua}]{cao2019unifying}
Yixin Cao, Xiang Wang, Xiangnan He, Zikun Hu, and Tat-Seng Chua. 2019.
\newblock Unifying knowledge graph learning and recommendation: Towards a
  better understanding of user preferences.
\newblock In \emph{The world wide web conference}, pages 151--161.

\bibitem[{Chen et~al.(2017)Chen, Tian, Yang, and
  Zaniolo}]{chen2017multilingual}
Muhao Chen, Yingtao Tian, Mohan Yang, and Carlo Zaniolo. 2017.
\newblock Multilingual knowledge graph embeddings for cross-lingual knowledge
  alignment.
\newblock In \emph{Proceedings of the 26th International Joint Conference on
  Artificial Intelligence}, pages 1511--1517.

\bibitem[{Jin et~al.(2021)Jin, Yuan, Xiong, Yu, Tan, Chen, Huang, Liu, and
  Yu}]{jin2021biomedical}
Qiao Jin, Zheng Yuan, Guangzhi Xiong, Qianlan Yu, Chuanqi Tan, Mosha Chen,
  Songfang Huang, Xiaozhong Liu, and Sheng Yu. 2021.
\newblock Biomedical question answering: A comprehensive review.
\newblock \emph{arXiv preprint arXiv:2102.05281}.

\bibitem[{Kipf and Welling(2016)}]{kipf2016semi}
Thomas~N Kipf and Max Welling. 2016.
\newblock Semi-supervised classification with graph convolutional networks.
\newblock \emph{arXiv preprint arXiv:1609.02907}.

\bibitem[{Kuhn(1955)}]{kuhn1955hungarian}
Harold~W Kuhn. 1955.
\newblock The hungarian method for the assignment problem.
\newblock \emph{Naval research logistics quarterly}, 2(1‐2):83--97.

\bibitem[{Lin et~al.(2021)Lin, Song, Luo et~al.}]{lin2021echoea}
Xueyuan Lin, Wenyu Song, Haoran Luo, et~al. 2021.
\newblock Echoea: Echo information between entities and relations for entity
  alignment.
\newblock \emph{arXiv preprint arXiv:2107.03054}.

\bibitem[{Lindberg et~al.(1993)Lindberg, Humphreys, and
  McCray}]{lindberg1993unified}
Donald~AB Lindberg, Betsy~L Humphreys, and Alexa~T McCray. 1993.
\newblock The unified medical language system.
\newblock \emph{Yearbook of Medical Informatics}, 2(01):41--51.

\bibitem[{Liu et~al.(2021)Liu, Hong, Wang, Chen, Kharlamov, Dong, and
  Tang}]{liu2021self}
Xiao Liu, Haoyun Hong, Xinghao Wang, Zeyi Chen, Evgeny Kharlamov, Yuxiao Dong,
  and Jie Tang. 2021.
\newblock A self-supervised method for entity alignment.
\newblock \emph{arXiv preprint arXiv:2106.09395}.

\bibitem[{Mao et~al.(2020)Mao, Wang, Xu, Wu, and Lan}]{mao2020relational}
Xin Mao, Wenting Wang, Huimin Xu, Yuanbin Wu, and Man Lan. 2020.
\newblock Relational reflection entity alignment.
\newblock In \emph{Proceedings of the 29th ACM International Conference on
  Information \& Knowledge Management}, pages 1095--1104.

\bibitem[{Nguyen et~al.(2020)Nguyen, Huynh, Yin, Van~Tong, Sakong, Zheng, and
  Nguyen}]{nguyen2020entity}
Tam~Thanh Nguyen, Thanh~Trung Huynh, Hongzhi Yin, Vinh Van~Tong, Darnbi Sakong,
  Bolong Zheng, and Quoc Viet~Hung Nguyen. 2020.
\newblock Entity alignment for knowledge graphs with multi-order convolutional
  networks.
\newblock \emph{IEEE Transactions on Knowledge and Data Engineering}.

\bibitem[{Pennington et~al.(2014)Pennington, Socher, and
  Manning}]{pennington2014glove}
Jeffrey Pennington, Richard Socher, and Christopher~D Manning. 2014.
\newblock Glove: Global vectors for word representation.
\newblock In \emph{Proceedings of the 2014 conference on empirical methods in
  natural language processing (EMNLP)}, pages 1532--1543.

\bibitem[{Savenkov and Agichtein(2016)}]{savenkov2016crqa}
Denis Savenkov and Eugene Agichtein. 2016.
\newblock Crqa: Crowd-powered real-time automatic question answering system.
\newblock In \emph{Fourth AAAI conference on human computation and
  crowdsourcing}.

\bibitem[{Sun et~al.(2021)Sun, Chen, and Hu}]{sun2021knowing}
Zequn Sun, Muhao Chen, and Wei Hu. 2021.
\newblock Knowing the no-match: Entity alignment with dangling cases.

\bibitem[{Sun et~al.(2017)Sun, Hu, and Li}]{sun2017cross}
Zequn Sun, Wei Hu, and Chengkai Li. 2017.
\newblock Cross-lingual entity alignment via joint attribute-preserving
  embedding.
\newblock In \emph{International Semantic Web Conference}, pages 628--644.
  Springer.

\bibitem[{Sun et~al.(2018)Sun, Hu, Zhang, and Qu}]{sun2018bootstrapping}
Zequn Sun, Wei Hu, Qingheng Zhang, and Yuzhong Qu. 2018.
\newblock Bootstrapping entity alignment with knowledge graph embedding.
\newblock In \emph{Proceedings of the 27th International Joint Conference on
  Artificial Intelligence}, pages 4396--4402.

\bibitem[{Trisedya et~al.(2019)Trisedya, Qi, and Zhang}]{trisedya2019entity}
Bayu~Distiawan Trisedya, Jianzhong Qi, and Rui Zhang. 2019.
\newblock Entity alignment between knowledge graphs using attribute embeddings.
\newblock In \emph{Proceedings of the AAAI Conference on Artificial
  Intelligence}, volume~33, pages 297--304.

\bibitem[{Veli{\v{c}}kovi{\'c} et~al.(2017)Veli{\v{c}}kovi{\'c}, Cucurull,
  Casanova, Romero, Lio, and Bengio}]{velivckovic2017graph}
Petar Veli{\v{c}}kovi{\'c}, Guillem Cucurull, Arantxa Casanova, Adriana Romero,
  Pietro Lio, and Yoshua Bengio. 2017.
\newblock Graph attention networks.
\newblock \emph{arXiv preprint arXiv:1710.10903}.

\bibitem[{Wang et~al.(2018)Wang, Lv, Lan, and Zhang}]{wang2018cross}
Zhichun Wang, Qingsong Lv, Xiaohan Lan, and Yu~Zhang. 2018.
\newblock Cross-lingual knowledge graph alignment via graph convolutional
  networks.
\newblock In \emph{Proceedings of the 2018 Conference on Empirical Methods in
  Natural Language Processing}, pages 349--357.

\bibitem[{Wu et~al.(2019)Wu, Liu, Feng, Wang, Yan, and Zhao}]{wu2019relation}
Yuting Wu, Xiao Liu, Yansong Feng, Zheng Wang, Rui Yan, and Dongyan Zhao. 2019.
\newblock Relation-aware entity alignment for heterogeneous knowledge graphs.
\newblock \emph{arXiv preprint arXiv:1908.08210}.

\bibitem[{Xiong et~al.(2017)Xiong, Power, and Callan}]{xiong2017explicit}
Chenyan Xiong, Russell Power, and Jamie Callan. 2017.
\newblock Explicit semantic ranking for academic search via knowledge graph
  embedding.
\newblock In \emph{Proceedings of the 26th international conference on world
  wide web}, pages 1271--1279.

\bibitem[{Xu et~al.(2020)Xu, Song, Feng, Song, and Yu}]{xu2020coordinated}
Kun Xu, Linfeng Song, Yansong Feng, Yan Song, and Dong Yu. 2020.
\newblock Coordinated reasoning for cross-lingual knowledge graph alignment.
\newblock In \emph{Proceedings of the AAAI Conference on Artificial
  Intelligence}, volume~34, pages 9354--9361.

\bibitem[{Xu et~al.(2019)Xu, Wang, Yu, Feng, Song, Wang, and Yu}]{xu2019cross}
Kun Xu, Liwei Wang, Mo~Yu, Yansong Feng, Yan Song, Zhiguo Wang, and Dong Yu.
  2019.
\newblock Cross-lingual knowledge graph alignment via graph matching neural
  network.
\newblock In \emph{Proceedings of the 57th Annual Meeting of the Association
  for Computational Linguistics}, pages 3156--3161.

\bibitem[{Yu et~al.(2017)Yu, Yin, Hasan, dos Santos, Xiang, and
  Zhou}]{yu2017improved}
Mo~Yu, Wenpeng Yin, Kazi~Saidul Hasan, Cicero dos Santos, Bing Xiang, and Bowen
  Zhou. 2017.
\newblock Improved neural relation detection for knowledge base question
  answering.
\newblock In \emph{Proceedings of the 55th Annual Meeting of the Association
  for Computational Linguistics (Volume 1: Long Papers)}, pages 571--581.

\bibitem[{Yuan et~al.(2020)Yuan, Zhao, Sun, Li, Wang, and Yu}]{yuan2020coder}
Zheng Yuan, Zhengyun Zhao, Haixia Sun, Jiao Li, Fei Wang, and Sheng Yu. 2020.
\newblock Coder: Knowledge infused cross-lingual medical term embedding for
  term normalization.
\newblock \emph{arXiv preprint arXiv:2011.02947}.

\bibitem[{Zeng et~al.(2021)Zeng, Zhao, Tang, Li, Luo, and
  Zheng}]{zeng2021towards}
Weixin Zeng, Xiang Zhao, Jiuyang Tang, Xinyi Li, Minnan Luo, and Qinghua Zheng.
  2021.
\newblock Towards entity alignment in the open world: An unsupervised approach.
\newblock \emph{arXiv preprint arXiv:2101.10535}.

\bibitem[{Zeng et~al.(2020)Zeng, Zhao, Tang, and Lin}]{zeng2020collective}
Weixin Zeng, Xiang Zhao, Jiuyang Tang, and Xuemin Lin. 2020.
\newblock Collective entity alignment via adaptive features.
\newblock In \emph{2020 IEEE 36th International Conference on Data Engineering
  (ICDE)}, pages 1870--1873. IEEE.

\bibitem[{Zhao et~al.(2020)Zhao, Zeng, Tang, Wang, and
  Suchanek}]{zhao2020experimental}
Xiang Zhao, Weixin Zeng, Jiuyang Tang, Wei Wang, and Fabian Suchanek. 2020.
\newblock An experimental study of state-of-the-art entity alignment
  approaches.
\newblock \emph{IEEE Transactions on Knowledge \& Data Engineering}, (01):1--1.

\bibitem[{Zhu et~al.(2021{\natexlab{a}})Zhu, Ma, and Wang}]{zhu2021raga}
Renbo Zhu, Meng Ma, and Ping Wang. 2021{\natexlab{a}}.
\newblock Raga: Relation-aware graph attention networks for global entity
  alignment.
\newblock In \emph{PAKDD (1)}, pages 501--513. Springer.

\bibitem[{Zhu et~al.(2021{\natexlab{b}})Zhu, Liu, Wu, and Du}]{zhu2021relation}
Yao Zhu, Hongzhi Liu, Zhonghai Wu, and Yingpeng Du. 2021{\natexlab{b}}.
\newblock Relation-aware neighborhood matching model for entity alignment.
\newblock In \emph{Proceedings of the AAAI Conference on Artificial
  Intelligence}, volume~35, pages 4749--4756.

\end{thebibliography}
\bibliographystyle{acl_natbib}

\appendix

\section{Appendix}
\subsection{Practiacl Evaluation Results}
\label{sec:appendixA1}
For completeness, this appendix reports the EA results on DBP15K in practiacl evaluation setting (Table~\ref{tab:Table 6}).We compared our methods with the RAGA, since we adopt the part of graph embedding in RAGA in our framework. 
\begin{table}[h]
  \setlength\tabcolsep{0.8pt}
  \small
  \centering
    \begin{tabular}{lcrrcrrcrr}
    \toprule
     &
      \multicolumn{3}{c}{ZH-EN} &
      \multicolumn{3}{c}{JA-EN} &
      \multicolumn{3}{c}{FR-EN}
      \\
     &
      \multicolumn{1}{l}{@1} &
      \multicolumn{1}{l}{@10} &
      \multicolumn{1}{l}{MRR} &
      \multicolumn{1}{l}{@1} &
      \multicolumn{1}{l}{@10} &
      \multicolumn{1}{l}{MRR} &
      \multicolumn{1}{l}{@1} &
      \multicolumn{1}{l}{@10} &
      \multicolumn{1}{l}{MRR}
      \\
    \midrule
    \textbf{local } &
       &
       &
       &
       &
       &
       &
       &
       &
      
      \\
    \midrule
    Init-Emb &
      .570  &
      \multicolumn{1}{c}{.686 } &
      \multicolumn{1}{c}{.611 } &
      .633  &
      \multicolumn{1}{c}{.753 } &
      \multicolumn{1}{c}{.676 } &
      .807  &
      \multicolumn{1}{c}{.890 } &
      \multicolumn{1}{c}{.835 }
      \\
    RAGA &
      .725  &
      \multicolumn{1}{c}{.903 } &
      \multicolumn{1}{c}{.790 } &
      .773  &
      \multicolumn{1}{c}{.931 } &
      \multicolumn{1}{c}{.829 } &
      .884  &
      \multicolumn{1}{c}{.972 } &
      \multicolumn{1}{c}{.917 }
      \\
    UED &
      .751  &
      \multicolumn{1}{c}{.892 } &
      \multicolumn{1}{c}{.802 } &
      .793  &
      \multicolumn{1}{c}{.918 } &
      \multicolumn{1}{c}{.839 } &
      .911  &
      \multicolumn{1}{c}{.974 } &
      \multicolumn{1}{c}{.934 }
      \\
    \midrule
    \textbf{global } &
       &
       &
       &
       &
       &
       &
       &
       &
      
      \\
    \midrule
    RAGA &
      .834  &
       &
       &
      .742  &
       &
       &
      .929  &
       &
      
      \\
    UED(DAA) &
      .799  &
       &
       &
      .769  &
       &
       &
      .935  &
       &
      
      \\
    UED &
      \textbf{.847}  &
       &
       &
      \textbf{.890}  &
       &
       &
      \textbf{.966}  &
       &
      
      \\
    \bottomrule
    \end{tabular}%
  \caption{EA results on DBP15K (practical setting). @1 and @10 denotes the Hits@1 and Hits@10. The UED(DAA) refer to the variant of UED by replacing the OTP with DAA.}
  \label{tab:Table 6}%
\end{table}%

\end{document}